# A Lifetime Extended Energy Management Strategy for Fuel Cell Hybrid Electric Vehicles via Self-Learning Fuzzy Reinforcement Learning

Liang GUO, *Student Member, IEEE*, Zhongliang LI *Member, IEEE*, and Rachid OUTBIB

*Abstract*— Modeling difficulty, time-varying model, and uncertain external inputs are the main challenges for energy management of fuel cell hybrid electric vehicles. In the paper, a fuzzy reinforcement learning-based energy management strategy for fuel cell hybrid electric vehicles is proposed to reduce fuel consumption, maintain the batteries' long-term operation, and extend the lifetime of the fuel cells system. Fuzzy Q-learning is a model-free reinforcement learning that can learn itself by interacting with the environment, so there is no need for modeling the fuel cells system. In addition, frequent startup of the fuel cells will reduce the remaining useful life of the fuel cells system. The proposed method suppresses frequent fuel cells startup by considering the penalty for the times of fuel cell startups in the reward of reinforcement learning. Moreover, applying fuzzy logic to approximate the value function in Q-Learning can solve continuous state and action space problems. Finally, a python-based training and testing platform verify the effectiveness and self-learning improvement of the proposed method under conditions of initial state change, model change and driving condition change.

## I. INTRODUCTION

Nowadays, fossil energy anxiety and climate problems by vehicle pollution are increasingly diverting more public's attention from internal combustion engine (ICE) vehicles to renewable energy vehicles. Fuel cell hybrid electric vehicles (FCHEV) have the advantages of no pollution, fast charging and high efficiency. The proton exchange membrane fuel cell (PEMFC) is an electrochemical device that only produces water with high-purity hydrogen and air, and has the characteristics of high energy density and high efficiency. The fuel cells (FCs) usually have slow dynamic performance, so an auxiliary energy storage system is needed to absorb the shock when the load power changes rapidly.

Energy management strategy (EMS) is the power allocation strategy between different energy sources for the hybrid energy system. For the FCHEV, a well-designed EMS can help to reduce fuel consumption and maintain the batteries' long time operation. Nevertheless, the difficulty for accurate modeling of FCs, time-varying model due to the degradation, and uncertain external driving conditions are challenging difficulties for the EMS problem of FCHEV. Rule-based EMS, such as traditional fuzzy logic-based EMS [1], is highly dependent on experience, hardly adapts to unknown future operating conditions or model changes, and cannot reach optimal results. The optimal-based EMS can obtain the optimal solution, such as Pontryagin's minimum principle (PMP) [2]. However, optimal-based methods are highly sensitive to modeling accuracy, so it is difficult to adapt to the model changes caused by degradation, and also needs prior knowledge of future operating conditions

Reinforcement learning (RL), a kind of machine learning, is increasingly applied to solve the optimal problem of energy management [3] because of its self-learning and model-free characteristics. RL-based EMS can learn itself from the interaction with the environment, so there is no need to detail the model of the controlled system or know the prior information on driving conditions. A Q-Learning-based EMS for fuel cell hybrid electric vehicles is proposed in [4], but is hard to deal with continuous space problems for high-dimensional computation. Deep Q-network (DQN) is proposed via deep neural networks (DNNs) to approximate the value function [5], so that the continuous state space problems can be solved. Deep deterministic policy gradient (DDPG) [6] approximates both the value function and policy function by DNNs, so that it can deal with continuous state and action.

For the fuel cells system, [7] shows that start/stop, idling, load changing, and high power load those factors are the main causes of fuel cell degradation. Among them, frequent startup has a major impact on reducing the lifespan of fuel cells[8]. Consider the degradation influence factor in the reward function of RL, so that RL-based EMS can learn to avoid behaviors that major cause FCs degradation. A DQN-based EMS with prioritized experience replay is proposed in [9], to extend the fuel cells lifetime of the FCHEV. However, the difficulty of tuning parameters and the computation burden of those deep reinforcement learning methods limit their applications for real-time online learning.

Applied fuzzy logic to approximate the value function of RL, fuzzy Q-learning (FQL) is proposed in [10], which is the first fuzzy reinforcement learning (FRL). An FQL-based EMS for a hybrid electric vehicle is proposed in [11], in which actions are obtained by the fuzzy logic controller, and Q-function is approximated with BP neural network to tune the parameters, which does not take advantage of the

This work has been supported by the ANR DEAL (contract ANR-20-CE05-0016-01). This work has also been partially funded by the CNRS Energy unit (Cellule Energie) through the project "PEPS GIALE".

Liang GUO is with the Laboratory LIS (UMR CNRS 7020) of Aix-Marseille University, Marseille, 13397 Cedex 20, France (e-mail: liang.guo@lis-lab.fr).

Zhongliang LI is with the Laboratory LIS (UMR CNRS 7020) of Aix-Marseille University, Marseille, 13397 Cedex 20, France (e-mail: zhongliang.li@lis-lab.fr).

Rachid OUTBIB is with the Laboratory LIS (UMR CNRS 7020) of Aix-Marseille University, Marseille, 13397 Cedex 20, France (e-mail: rachid.outbib@lis-lab.fr).

function approximation and generalization of fuzzy logic. A fuzzy inference system (FIS) for reinforcement learning was proposed in [12], which elaborated the architecture of the approximation method for the value function and policy function to solve continuous space problems. A fuzzy rule value reinforcement learning (FQL) based EMS is proposed for fuel cell hybrid electric vehicles [14], which approximates the rule value with FIS by RL.

In the paper, an FQL-based EMS for the FCHEV is proposed to extend the lifetime of the fuel cell system by involving the degradation of the objective function. By combining with the FIS of fuzzy logic, the state-action value function is approximated, and the operation of continuous states and actions is realized. The agent of the proposed method performs model-free self-learning by interacting with the environment, so that the proposed EMS controller does not need the controlled system model, and can also adapt to the time-varying model and uncertain working conditions.

## II. Modeling of the Energy System

For the energy system of FCHEV, to absorb the power fluctuation of the load, a battery system needs to be added, which is due to the slow dynamic response of the fuel cells, and can make the fuel cells work in high-efficiency points. The studied energy system of the FCHEV is as shown in Fig. 1. The fuel cell system and the battery system are connected to the DC bus through DC/DC converters to supply power to the load motor system or recover braking or deceleration energy from the motor system. The control system is composed of high-level and low-level controllers, and the EMS studied in the paper is the high-level controller. The specific models of the energy system will be analyzed in this section.

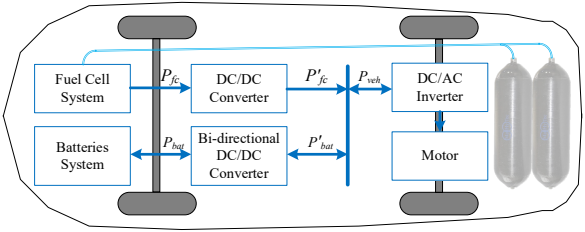

Fig. 1. Energy system for fuel cell hybrid electric vehicle.

### A. Vehicle dynamics model

The mechanics and power model of the vehicle are shown as (1) with the velocity $v$ and the road slope $\theta$.

$$\begin{cases} F_m = \frac{1}{2}C_D A \rho v^2 + Gf\cos\theta + G\sin\theta + m\frac{dv}{dt} \\ P_{veh} = F_m \cdot v / \eta_m \end{cases} \quad (1)$$

where $F_m$ represents the driving force provided by the motor, which equals the sum of the air resistance, the rolling resistance, the slope resistance and the acceleration force. $\rho$ and $C_D$ represent air density and air resistance coefficient respectively. $A$ represents the windward surface volume of the vehicle body, and $v$ represents the vehicle velocity. $m$ represents the vehicle mass. $G = mg$ represents the gravity of the vehicle, and $f$ represents the sliding resistance coefficient. $P_{veh}$ represents the required power of the motor, $\eta_m$ represents the transmission efficiency of the motor.

For the studied vehicle, the vehicle weight is $2500\ kg$, the windward area is $1.8\ m^2$, the air density is $1.25\ kg/m^2$, the air resistance coefficient is 0.3, the rolling friction coefficient is 0.01, and the total mechanical transmission efficiency is set as 90%, the gravity acceleration is 9.8 $m/s^2$.

### B. Fuel cells system model

The voltage model of a single cell of the fuel cell can be expressed as follows:

$$V_{cell} = E_0 + \frac{\Delta TS}{nF} - \frac{RT}{nF}\ln\left(\frac{P_{H_2O}}{P_{H_2}\sqrt{P_{O_2}}}\right) \\ -\frac{RT}{\alpha F}\ln\left(\frac{i_{fc}+i_{loss}}{i_0}\right) - \frac{RT}{nF}\ln\left(\frac{I_{lim}}{I_{lim}-i_{fc}}\right) - i_{fc}R_{ohm} \quad (2)$$

where $E_0 = 1.23\ V$ is the open-circuit voltage of fuel cell reaction at standard atmospheric pressure, $R = 8.3145$ is the gas constant, $T = 333.15\ K$ is the fuel cell temperature, $\Delta T = T - 273.15$, $n = 2$, $F = 96485$ is Faraday constant, $\alpha = 1$ is the transfer coefficient, $P$ is the local pressure of the reactants and products at this atmospheric pressure. $i_{fc}$ is the current density. $i_{loss} = 2mA/cm^2$ is the current loss, $i_0 = 0.003mA/cm^2$ is the exchange current density. $I_{lim} = 1.6A/cm^2$ is the limiting current density. $R_{ohm}$ is the fuel cell resistance.

For the FC stack, the model is as follows:

$$\begin{aligned} V_{fc} &= n_{cell} \cdot V_{cell} \\ I_{fc} &= A_{fc} \cdot i_{fc} \end{aligned} \quad (3)$$

where $n_{cell}$ is the number of single FCs, and $A_{fc}$ is the active area of the FC electrode plate. Then the hydrogen consumption model of the FC stack can be derived as follows:

$$\dot{m}_{H_2} = M_{H_2}\frac{I_{fc}}{nF} = \frac{M_{H_2}P_{fc}}{nV_{fc}F} \quad (4)$$

where $\dot{m}_{H_2}$ is the rate at which hydrogen is consumed, and $M_{H_2}$ is the molar mass of hydrogen. $P_{fc}$ is the output power of FCs. The converter model will only be concerned about its power characteristics. The DC/DC converter power efficiency model for FCs is as follows:

$$P_{fc} = P_{fc}' / \eta_{dc}(P_{fc}') + P_{aux} \quad (5)$$

where $P_{fc}'$ is the output power of the FC system. It is considered that $P_{fc}'$ is equal to the power command from the control strategy. $\eta_{dc}$ is the efficiency of DC/DC converter for fuel cells. $P_{aux}$ is the auxiliary system, and it can be considered as a constant current load $I_{aux} = 2.0\ A$.

The fuel cells parameters are $n_{cell} = 200$, the effective area of the electrode is $A_{fc} = 324\ cm^2$, the pressure of anode hydrogen is 50 kPa over atmosphere pressure, and cathode oxygen is obtained from the air by natural aspiration. As shown in Fig. 2, when the current is 437 A, the FC power reaches the max power of 104 kW and the efficiency is 43.19 %; When the current is 63.2 A, the FC efficiency reaches the max efficiency of 54.49 %, and the power is 15.7 kW.

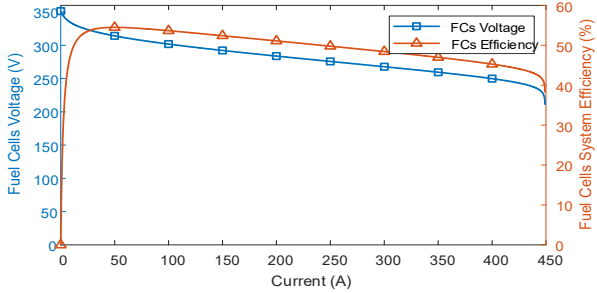

Fig. 2. The output voltage and efficiency of fuel cells

### C. Batteries system model

The battery is modeled using a simple one-order circuit model [15]. The output current of the battery and the evolution of the battery SOC are characterized by the following:

$$I_{bat} = \frac{V_{oc} - \sqrt{V_{oc}^2 - 4R_{bat}P_{bat}}}{2R_{bat}} \quad (6)$$

$$SOC_{bat} = SOC_{bat}(0) - \int_0^t I_{bat}/Q_{bat} dt$$

where $I_{bat}$ is the output current of the batteries. When $I_{bat} > 0$, the battery is discharged, and when $I_{bat} < 0$, the battery is charged. $SOC_{bat}$ is the state of charge (SOC) of batteries. $Q_{bat}$ is the battery capacity. Especially, the open-circuit voltage $V_{oc}$ and the internal resistance $R_{bat}$ are dependent on $SOC_{bat}$, which are shown in Fig. 3.

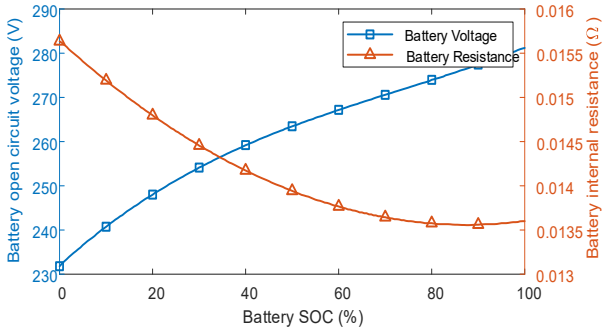

Fig. 3. The characteristics of the batteries

Considering the power loss of the DC/DC converter, the battery output power is expressed as:

$$P_{bat} = \begin{cases} P_{bat}'/\eta_{bdc} & (P_{bat}' > 0) \\ P_{bat}' \cdot \eta_{bdc} & (P_{bat}' < 0) \end{cases} \quad (7)$$

where $P_{bat}'$ is the output power of the power converter whose efficiency is $\eta_{abc}$. According to the power balance, the power of the batteries system is determined by the power balance:

$$P_{bat}' = P_{veh} - P_{fc}' \quad (8)$$

In the paper, the total capacity of the studied batteries is set as 6.6 Ah, and the standard voltage is 244.8V.

## III. FUZZY Q-LEARNING PRINCIPLE

### A. EMS problem formulation

The main task of the EMS in the paper is to maximize the objective function by adjusting the power distribution among different energy sources. The objective function consists of two parts. The first part is the instantaneous reward, which takes into account the fuel consumption rate and the deviation of the SOC of the battery from the reference value. The second part is the episode reward, which considers the number of fuel cells starts when in the terminal state. The objective function $J$ of one episode is formulated mathematically as follows:

$$\max J = \int_0^T r(t)dt + k_{start}N_{start}$$
$$r(t) = -\dot{m}_{H_2}(t) - k_{bat}(SOC_{bat}(t) - SOC_{ref})^2 \quad (9)$$

where $r(t)$ is the instantaneous reward, $\dot{m}_{H_2}$ is the hydrogen consumption rate, $SOC_{bat}$ is the SOC of batteries, $SOC_{ref}$ is the preset reference of SOC corresponding to the battery characteristics, $N_{start}$ is the number of fuel cell starts. In the paper, the EMS is dedicated to determining $P_{fc}(t), t \in [0,T]$ using the observables $[P_{veh}(t), SOC_{bat}(t)]$ to achieve the maximization of the objective function $J$.

### B. Fuzzy Interrace System for the Energy Management Problem of FCHEV

Fuzzy logic is a mathematical language that imitates the human brain's uncertain concept judgment and reasoning thinking. A basic fuzzy inference system (FIS) consists of four parts: fuzzifier, defuzzifier, inference engine, and knowledge base.

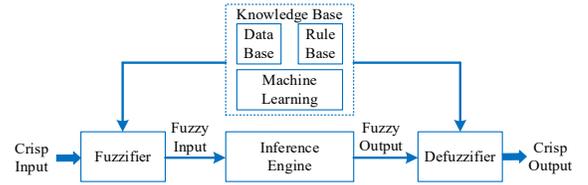

Fig. 4. Fuzzy Interrace System scheme

The control based on FIS is an effective and widely used method to deal with energy management problems. In our application, the EMS deals with a multi-input single-output FIS control system. As shown in Fig. 4, the crisp input is the system state $s = [P_{veh}, SOC_{bat}]$, and the crisp output is the action of the control system $a = [P_{fc}]$.

With the fuzzifier, the fuzzy state $\phi(s) = [\phi_1, \phi_2, ..., \phi_m]$ of the energy system can be derived by predefined membership functions, and $\phi_i(s)$ represents the fired strength of the $i^{th}$ rule. The meaning of the fuzzy sets ["NH", "NM", "NL", "ZO", "PL", "PM", "PH"] for $P_{veh}$ are "Negative High", "Negative Middle", "Negative Low",

"Zero", "Positive Low", "Positive Middle", and "Positive High". For another state $SOC_{bat}$, the meaning of ["VL", "L", "M", "H", "VH"] are "Very Low", "Low", "Middle", "High", and "Very High". The fuzzy sets of the two input states are shown in Table I and Table II. Then the membership functions are chosen as shown in Fig. 5.

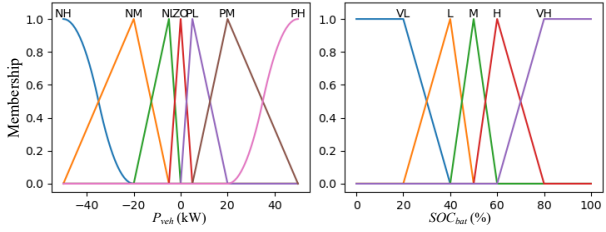

(a) Membership of $P_{veh}$  (b) Membership of $SOC_{bat}$

Fig. 5. Membership functions of state variables

TABLE I. VEHICLE REQUIRED POWER FUZZY TABLE

| $P_{veh}$ Fuzzy sets | NH | NM | NL | ZO | PL | PM | PH |
|---|---|---|---|---|---|---|---|
| Typical Value (kW) | -50 | -20 | -10 | 0 | 10 | 20 | 50 |

TABLE II. BATTERIES SYSTEM POWER FUZZY TABLE

| $SOC_{bat}$ Fuzzy sets | VL | L | M | H | VH |
|---|---|---|---|---|---|
| Typical Value (%) | 20 | 40 | 50 | 60 | 80 |

The membership functions of the two crisp input states can then be transformed into fuzzy states $\phi(s)$ with fuzzy logic operation "AND". The number of states in $\phi(s)$ is identical to the number of rules. In our case, the rule is set for each combination of the two fuzzy sets. Hence, the dimensional number of fuzzy state $\phi(s)$ is $M = 35$.

Traditionally, fuzzy rules can be constructed using experienced data and/or engineering experience. The logic rules are formed like:

**IF** $P_{veh}$ is " Positive High" (PH), **AND** $SOC_{bat}$ is "Very Low"(VL), **THEN** $P_{fc}$ is "Super High" (SH).

The inference engine deduces, then the fuzzy output based on each rule. The control action is calculated by deffuzier combining all fuzzy outputs. For instance, the calculation can be realized using the weighted average defuzzification method as:

$$y = \frac{\sum_{i=1}^{M} y_i \phi_i(s)}{\sum_{i=1}^{M} \phi_i(s)} \quad (10)$$

where $y_i$ is the $i^{th}$ fuzzy output, and $y$ is the defuzzied value of $y_i$ by the weighted average method. The output action and the function approximator in the paper can be cauletd by this method.

TABLE III. FUEL CELLS SYSTEM POWER FUZZY TABLE

| $P_{fc}$ Fuzzy Sets | ZO | SL | VL | L | M | H | VL | SH |
|---|---|---|---|---|---|---|---|---|
| Typical Value (kW) | 0 | 1 | 2 | 5 | 10 | 20 | 50 | 100 |

For the action of $P_{fc}$, the typical values of the fuzzy output sets are assembled as $U = \{U_1, U_2, ..., U_N\}$ in this study are shown in 0 The dimension of output fuzzy sets is set as $N = 8$. Here, a non-equidistant method is used for the division of fuzzy sets, and the state with lower power is divided more finely.

## C. Reinforcement learning principles

Reinforcement learning realizes continuous self-learning by interacting with the environment, so it is possible to break the upper limit of human experience, or correct the error between the model and reality, to obtain a better optimal solution. Meanwhile, the environment is required to have Markov properties, which is the transition probability of the next state can be only determined by the current state $s(t)$ and the action $a(t)$. Thus, a sequence $[s(0), a(0), r(0), s(1), a(1), r(1), ...]$ can be obtained during the learning process until the terminal state. This process is a Markov decision process.

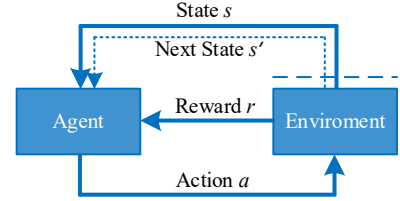

Fig. 6. Reinforcement Learning Principle

The goal of the RL agent is to maximize the cumulative rewards $J$ from the initial state $s_0$ to the terminal state $s_T$ at time $t = T$ by optimizing the policy.

$$J(\pi) = \mathbb{E}\left[\sum_{t=0}^{T} \gamma^t r(t)\right] \quad (11)$$

Where $\pi$ is the policy to obtain the action, and $\gamma$ is a discount factor, $0 \leq \gamma < 1$. $\gamma = 0$ means immediate return, $\gamma$ tends to 1 means future return. The cumulative discounted reward is used to evaluate the performance of the policy. However, the cumulative reward is difficult to calculate directly at each step. In Q-learning, time-difference method with a Q-function $Q(s, a)$ is used to replace the cumulative reward. The Q-value represents the state-action value at the time $t$, and it is shown in follows:

$$\begin{aligned} Q(s,a) &= \mathbb{E}_{s \sim \rho(s), a \sim \pi(s)}\left[\sum_{k=t}^{k=T} \gamma^{k-t} r(k)\right] \\ &= r(t) + \mathbb{E}_{s \sim \rho(s), a \sim \pi(s)}\left[\sum_{k=t+1}^{k=T} \gamma^{k-t} r(k)\right] \\ &= r(t) + \gamma \sum P(s'|s,a) Q(s', \pi(s')) \end{aligned} \quad (12)$$

where $s'$ is the next state, and $\pi(s')$ means all possible actions of the next state $s'$. $P(s'|s,a)$ means the probability of reaching the next state $s'$, under the conditions of state $s$ and action $a$. Therefore, how to estimate the expectation of the value function of the next state $s'$ is a key element of implementing Q-Learning. To achieve fast reinforcement, Q-learning adopts the maximization estimation method, to maximize the next Q-value $Q(s', a')$. Then the optimal Q function $Q^*(s, a)$ is shown in (13) with action $a$ when the state is $s$, which is also the well known Bellman equation.

$$\begin{aligned} Q^*(s,a) &= \mathbb{E}_{s \sim \rho(s), a \sim \pi(s)}\left[r + \gamma \max \mathbb{E}_{a' \sim \pi(s')}\left[Q(s',a')\right]\right] \\ &= \mathbb{E}_{s' \sim \rho(s)}\left[r + \gamma V(s')\right] \end{aligned} \quad (13)$$

where $V(s')$ is the value function of the next state $s'$. Then the optimal policy $\pi^*$ can be specified as follows:

$$\pi^*(s) = \arg\max_{a \sim \pi(s')} (Q^*(s,a)) \quad (14)$$

The update law of Q-Learning can be expressed as [16]:

$$Q(s,a) = Q(s,a) + \alpha[r + \gamma \max_{a' \sim \pi(s')} Q(s',a') - Q(s,a)] \quad (15)$$

where $\alpha \in (0,1)$ is the learning rate of Q-learning. When the learning rate is close to 1, it will be difficult to retain the past effective experience. When the learning rate is close to 0, the number of iterations will increase, which will bring a computational burden. According to general experience, $\alpha$ is set close to 0, and $\gamma$ is close to 1.

*D. Fuzzy Q-learning based EMS*

Since Q-Learning needs to be discretized when dealing with continuous states and actions space, and utilizes a large table to store state-action values Q-value, it will bring a serious computational burden with high-dimension. Using fuzzy membership functions as state approximators in Q-learning is an effective and faster way to solve continuous-state problems [17]. In FQL, $s$ is the crisp set of the input states which are converted into fuzzy states $\phi(s)$ with the membership functions of the FIS fuzzifier. Then the Q-value can be estimated by a fuzzy Q-function (16):

$$Q(\phi(s), a) = \frac{q^\top \phi(s)}{\sum \phi(s)} \quad (16)$$

where $Q(\phi(s), a)$ is the evaluated Q-function with defuzzifier, and $q = [q_{1,1}, q_{1,2}, \ldots, q_{M,N}]$ is a $[M \times N]$ q-array. $q_i$ is the $i^{th}$ q-array of the $i^{th}$ rule with the size of $[1 \times N]$, which contains the fuzzy values of each fuzzy action sets. Similarly, the value function of the next state can also be obtained by defuzzification:

$$V(\phi(s')) = \frac{v^\top \phi(s')}{\sum \phi(s')} \quad (17)$$

where $v = [v_1, v_2, \ldots, v_M]$ is the fuzzy state value of each rule at the next state $s'$, which evaluate by the maximization method.

$$v_i(\phi(s')) = \max_{a_i^* \in U} q_i(\phi_i(s'), a_i^*) \quad (18)$$

where $a_i^*$ is the optimal fuzzy action of the $i^{th}$ fuzzy next state $\phi_i(s')$, which is chosen from N fuzzy action of the set $U$.

$$a_i^*(s') = \arg\max_{a_i^* \in U} q_i(\phi_i(s'), a_i^*) \quad (19)$$

For $i$ th rule ($i \in \{1, \ldots, M\}$), the fuzzy rule can be expressed as:

**IF** *fuzzy state* $\phi_i(s)$, **THEN** *take* $a_i^*$ *with* $q_i$ *to update* $v_i$

And there will be $M$ fuzzy rules. The fuzzy Q-Learning will also be updated with the time-difference learning, and the target Q-function is set as: $Q_{target}(s,a) = r + \gamma V(s')$.

Then $\Delta q_i$ the increments of the fuzzy q-array corresponding to the $i^{th}$ rule are shown as:

$$\Delta q_i = [r + \gamma V(\phi(s')) - Q(\phi(s), a)] \frac{\phi_i(s)}{\sum_{i=1}^{M} \phi_i(s)} \quad (20)$$

where $\Delta q = [q_1, q_2, \ldots, q_M]$ is based on the fuzzifier with fuzzy state $\phi(s)$. And they are used in (21) to update the fuzzy q-arrays: $q_1$ and $q_2$ which are corresponding to each rule update:

$$q_i(\phi_i(s), a_i) := q_i(\phi_i(s), a_i) + \alpha \Delta q_i \quad (21)$$

where $a_i$ is the $i^{th}$ fuzzy action determined by the $\varepsilon - Greedy$ (22) to balance the exploration and exploitation of the FQL.

$$a_i = \begin{cases} \arg\max_{a_i \in U} q_i(\phi_i(s'), a_i) & \varepsilon \geq \mathcal{N}(0,1) \\ random \ a_i \in U & \varepsilon < \mathcal{N}(0,1) \end{cases} \quad (22)$$

where $\varepsilon \in [0,1]$ is the exploration rate of the FQL agent, and $\mathcal{N}(0,1)$ is a random value in the range of $[0,1]$. Finally, the actual action can be defuzzied with fuzzy action $a_i$ and fuzzy state $\phi(s')$.

$$a(s) = \frac{\sum_{i=1}^{M} a_i \phi_i(s)}{\sum_{i=1}^{M} \phi_i(s)} \quad (23)$$

With FQL, the continuous action and state space problems are solved with a fuzzy inference system, which can achieve a faster and smoother training, and it can adapt to the complex time-varying model through interaction with the environment. The pseudocode of the proposed FQL is shown in TABLE IV.

TABLE IV. THE PSEUDOCODE OF THE PROPOSED FQL

| Fuzzy Q-Learning (FQL) |
|---|
| Randomly initialize q-array $q$ with the size of $[M \times N]$ |
| $M$: the number of fuzzy rules; $N$: the number of fuzzy outputs sets $U$ |
| **for** $episode = 1$ **to** $L$ **do**: |
|   Reset the environment with the initialized state $s_0$ |
|   Obtain fuzzy state $\phi(s)$ with membership functions of each rule |
|   **for** $t = 1$ **to** $T$ **do**: |
|     Obtain fuzzy actions: $a_i$, with $\varepsilon - Greedy$ according to $q_i$ |
|     $a_i = \begin{cases} argmax q_i(\phi_i(s'), a_i) & \varepsilon \geq \mathcal{N}(0,1) \\ random \ a_i \in U & \varepsilon < \mathcal{N}(0,1) \end{cases}, i = 1, \ldots, M$ |
|     Take the actual action with defuzzifier: |
|     $a(s) = \frac{\sum_{i=1}^{M} a_i \phi_i(s)}{\sum_{i=1}^{M} \phi_i(s)}$ |
|     Observe the reward r and next state $s'$ |
|     Obtain the next fuzzy state $\phi(s')$, the fired strength of each rule |
|     Evaluate the fuzzy value of the next fuzzy state $\phi_i(s')$. |
|     $v_i(\phi_i(s')) = \max_{a_i^* \in U} q_i(\phi_i(s'), a_i^*)$ |
|     Get value funtion and Q-function with deffuzifier: |
|     $V(s') = \frac{\sum_{i=1}^{M} v_i(s')\phi_i(s')}{\sum_{i=1}^{M} \phi_i(s')}, Q = \frac{\sum_{i=1}^{M} q_i \phi_i(s)}{\sum_{i=1}^{M} \phi_i(s)}$ |
|     update the fuzzy q-array. |
|     $\Delta q_i = [r + \gamma V(\phi(s')) - Q(\phi(s), a)] \frac{\phi_i(s)}{\sum_{i=1}^{M} \phi_i(s)}$ |
|     $q_i(\phi_i(s), a_i) := q_{1,i}(\phi_i(s), a_i) + \alpha \Delta q_i$ |
|     Update state: $s \leftarrow s', \phi(s) \leftarrow \phi(s')$ |
|   **end for** |
| **end for** |

## IV. TRAINING AND TEST RESULTS ANALYSIS

A Python-based training and testing platform have been established for the proposed FQL-based EMS and the studied FCHEV. The processor is Intel(R) Core (TM) i5-9400H CPU @ 2.50GHz. The environment model of FCHEV is referred to the report of Toyota's *Mirai* FCHEV given by the Argonne National Laboratory. In this section, the training process and the test results of the proposed EMS are analyzed and discussed.

### A. Test driving cycles

The proposed EMS is tested using 2 standard driving cycles *Urban Dynamometer Driving Schedule* (UDDS) and *New European Driving Cycle* (NEDC). The velocity and power of the specific FCHEV under those 2 driving conditions are shown in Fig. 7. The proposed FQL-based EMS is trained only with "UDDS" and tested with both driving cycles.

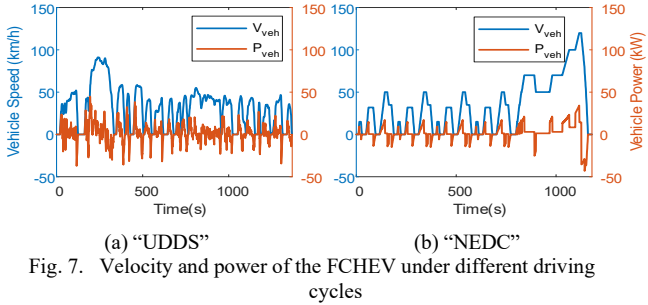

(a) "UDDS"  (b) "NEDC"
Fig. 7. Velocity and power of the FCHEV under different driving cycles

### B. Training and Test Results Analysis

To compare the performance of the proposed FQL with degradation and the FQL, we train the agents for both of them with the same parameters. The learning rate is set as $\alpha = 0.005$, and the decay rate is set $\gamma = 0.999$. The system states are constrained as $P_{veh}(t) \in [-50kW, 50kW]$ and $SOC_{bat}(t) \in [0\%, 100\%]$, and control action $P_{fc}(t) \in [0, 100\ kW]$. The initial state of $SOC_{bat}(t)$ is set as 50% during the training process. For the objective function in (9), $k_{SOC}$ =200, and $k_{start}$ =0.2. The exploration rate $\varepsilon$ decreases exponentially from 1 to 0.001. The total training episode is 1000, and the training time for the FQL-based EMS and the proposed FQL-based EMS with degradation is 15 and 17 minutes separately. The learning process of the proposed FQL-based EMS and the FQL-based EMS with degradation are shown in Fig. 8, which shows the average reward and the average hydrogen consumption of the proposed methods.

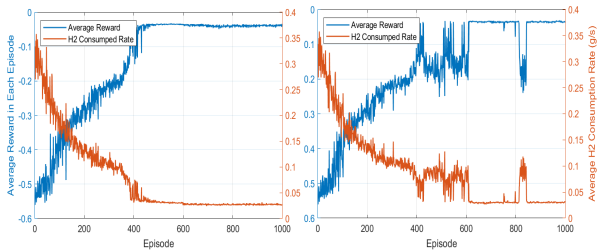

(a) without degradation factor  (b) with degradation factor
Fig. 8. The training process of the proposed FQL-based EMS for FCHEV.

Due to the tendency to random strategies in the early stage of training, there will be cases where the agent cannot fully go through a driving cycle and exits the current episode because the SOC reaches the boundary value. The average reward eventually tends to the highest value, and the average hydrogen consumption tends to the lowest value. Both EMS strategies achieve stable convergence.

In the test where the initial SOC value is 50% and the driving condition is "UDDS", the FQL-based EMS and the proposed FQL-based EMS with degradation are applied for testing. The SOC trajectories are shown in Fig. 9, and the actual $P_{fc}$ output of the EMS strategy are shown in Fig. 10. The SOC of the terminal state of the proposed FQL-based EMS with degradation is 48.95%, and the FQL's is 48.05%. For the control action of the EMS in the 2nd zoon figure of Fig. 10, the number of fuel cell starts is significantly reduced by the proposed method considering the degradation factor, which is better for prolonging the lifetime of the fuel cells.

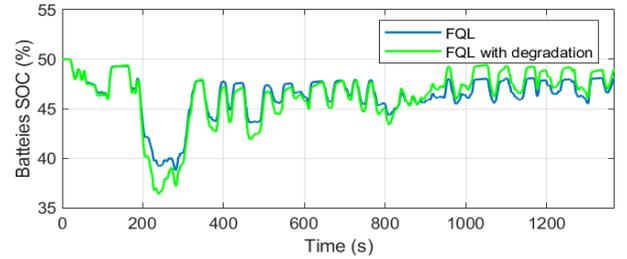

Fig. 9. The SOC trajectories of the FQL-based EMS and the of the proposed FQL-based EMS with degradation for FCHEV.

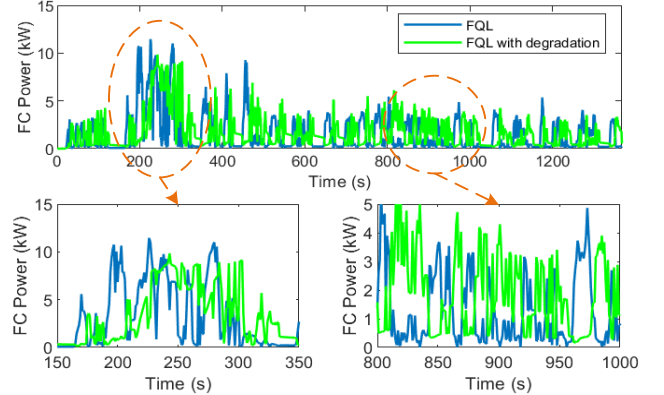

Fig. 10. The fuel cells system power of the FQL-based EMS and the of the proposed FQL-based EMS with degradation for FCHEV.

To verify the adaptability of the proposed method to the initial state, several tasks are being tested with the initial state of SOC 25%, 50% and 75%. To further verify the adaptability of the proposed method to the unknown driving condition, the tests with driving cycle "NEDC" are also carried out with the different initial states of SOC. Since the "UDDS" is the only driving condition used during the training process, "NEDC" is completely unknown to the well-trained agent. The SOC trajectories of the FQL-based EMS with degradation under driving cycles "UDDS" and "NEDC" are shown in Fig. 11. The test results verify the adaptability of the proposed method to changes in the initial state and driving conditions. Those are also the advantages of the model-free method, which is insensitive to model changes and does not require modeling for complex models. And the power allocation strategies of the proposed method with different initial SOC values are shown in Fig. 12. See detailed test results in TABLE V.

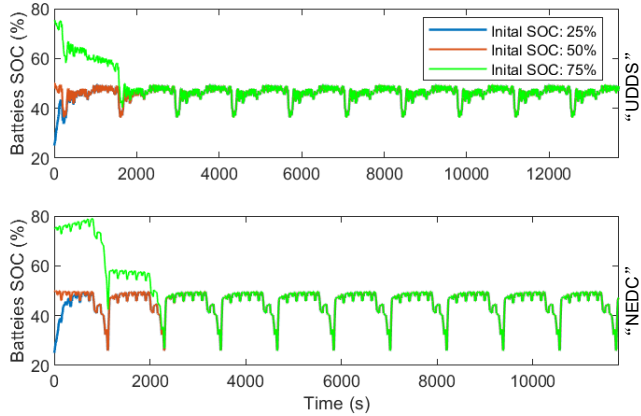

Fig. 11. The SOC trajectories of the proposed FQL-based EMS with degradation for FCHEV under the diving cycle "UDDS" and "NEDC"

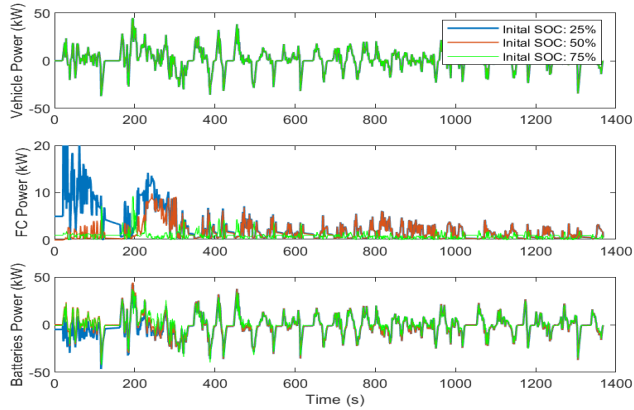

Fig. 12. The power allocation of the proposed FQL-based EMS with degradation for FCHEV under the driving cycle "UDDS"

TABLE V. TEST RESULTS AFTER 10 DRIVING CYCLES TIME OF THE PROPOSED FQL-BASED EMS WITH DEGRADATION

| Driving Cycle | Initial SOC | 1st Diving Cycle | | | 10th Driving Cycle | | |
|---|---|---|---|---|---|---|---|
| | | Average Reward | $H_2$ Rate (g/100km) | Final SOC | Average Reward | $H_2$ Rate (g/100km) | Final SOC |
| UDDS | 25% | -0.060 | 504.75 | 48.94% | -0.036 | 348.29 | 48.95% |
| | 50% | -0.035 | 339.70 | 48.95% | -0.036 | 348.29 | 48.95% |
| | 75% | -0.070 | 239.78 | 59.18% | -0.036 | 348.29 | 48.95% |
| NEDC | 25% | -0.070 | 520.40 | 47.16% | -0.044 | 361.02 | 47.16% |
| | 50% | -0.042 | 339.71 | 47.16% | -0.044 | 361.02 | 47.16% |
| | 75% | -0.167 | 233.76 | 58.07% | -0.044 | 361.02 | 47.16% |

## V. CONCLUSIONS

An FQL-based EMS with a degradation factor is proposed for the FCHEV, which combines the advantages of fuzzy logic and reinforcement learning. The FQL is a model-free RL method that can learn itself through interaction with the environment even the model of the environment is unknown. Also, a fuzzy inference system is used to approximate the Q-function for reinforcement learning to solve the continuous space problems and reduce computation significantly. The effectiveness of the proposed method is verified by python based training and test platform. While achieving the requirements of reducing fuel consumption and maintaining battery operation, the proposed method can also extend the lifetime of fuel cells by reducing the number of fuel cell starts. In addition, by changing the initial state and driving conditions in the test, the proposed EMS can still maintain good adaptability. Combining fuzzy reinforcement learning with the fuel cell degradation process helps to further enhance the potential of EMS for FCHEV.